\documentclass[11pt,a4paper]{article}

\usepackage[utf8]{inputenc}
\usepackage[T1]{fontenc}
\usepackage{lmodern}
\usepackage{microtype}
\usepackage[margin=1in]{geometry}
\usepackage{setspace}
\usepackage{graphicx}
\usepackage{xcolor}
\usepackage{url}
\usepackage[colorlinks=true,linkcolor=blue,citecolor=blue,urlcolor=blue]{hyperref}
\usepackage{amsmath,amssymb,amsthm}
\usepackage{booktabs,multirow,colortbl,array}
\usepackage{caption}
\usepackage{subcaption}
\usepackage{natbib}
\usepackage{authblk}

\definecolor{best}{RGB}{255,215,0}     
\definecolor{second}{RGB}{192,192,192} 
\definecolor{third}{RGB}{205,127,50}   
\newcommand{\best}[1]{\colorbox{best}{#1}}
\newcommand{\second}[1]{\colorbox{second}{#1}}
\newcommand{\third}[1]{\colorbox{third}{#1}}

\title{VL-MemKnG: Hybrid Memory with a Spatio-Temporal Knowledge Graph for Question Answering over Long Egocentric Navigation Trajectories}

\author[1,*]{Svetlana Lukina}
\author[1,*]{Mohamad Al Mdfaa}
\author[2]{Gloria Haro}
\author[3]{Sergey Zagoruyko}
\author[1]{Gonzalo Ferrer}

\affil[1]{Mobile Robotics Laboratory, Artificial Intelligence Center, Moscow, Russia}
\affil[2]{Intelligent Multimodal Vision Analysis Group, Department of Engineering, Universitat Pompeu Fabra, Barcelona, Spain}
\affil[3]{Independent Researcher}
\affil[*]{Correspondence: Svetlana Lukina, \texttt{svetlana.lukina@skoltech.ru}; 

Mohamad Al Mdfaa, \texttt{mohamad.almdfaa@skoltech.ru}}

\date{}

\begin{document}
\maketitle

\begin{abstract}
Answering navigation-relevant questions over long egocentric videos requires systems to retrieve and organize evidence distributed across distant temporal moments while maintaining spatial and contextual consistency. While full-context vision--language models can achieve strong answer quality, they are computationally expensive for long trajectories and inefficient for repeated querying. Recent graph-based approaches such as {\em VL-KnG} address this challenge through persistent spatio-temporal knowledge graphs with long-range object association and structured relational retrieval. However, graph-centric retrieval alone may underrepresent broader temporal continuity and contextual cues distributed across long video segments.
We present {\em VL-MemKnG}, a hybrid persistent memory framework that extends VL-KnG for temporally grounded video question answering by combining a spatio-temporal knowledge graph with persistent segment-level contextual memory. The knowledge graph preserves structured relational information and long-range object associations, while segment-level memory retains broader temporal and contextual continuity for long-horizon evidence retrieval. A hybrid retrieval-and-reasoning layer jointly operates over both memory representations to produce evidence-grounded answers and temporally organized supporting evidence.
We further introduce WalkieKnowledgeT+, an extension of the original WalkieKnowledge benchmark for long-horizon navigation-oriented video question answering over egocentric trajectories. WalkieKnowledgeT+ introduces temporally distributed reasoning tasks that require evidence aggregation across multiple non-cooccurring moments.
On WalkieKnowledgeT+, VL-MemKnG improves Top-1 retrieval accuracy from 58\% (best VL-KnG variant) to 67\% and increases Recall@1 from 34.50\% to 40.55\%, outperforming all compared methods, including Gemini 2.5 Pro and Qwen 3.5+. These improvements arise from combining knowledge graph-based memory with segment-level contextual memory that preserves broader temporal continuity, which is particularly beneficial for temporal-global and temporally scattered aggregation questions requiring evidence retrieval across multiple non-contiguous temporal intervals. Despite improved retrieval quality, VL-MemKnG maintains efficient query-time inference through persistent precomputed memory representations.

\vspace{0.5em}
\noindent\textbf{Keywords:}
Knowledge Graphs;
Embodied Scene Understanding;
Vision-Language Models;
Egocentric Video;
Long-horizon Video Understanding;
Retrieval-Augmented Generation;
Video Question Answering;
Spatio-Temporal Reasoning.

\end{abstract}

\section{Introduction}

Robots navigating real-world environments do not observe isolated images. Instead, they move through space and accumulate a long egocentric video trajectory that records their navigation experience over time. As the robot explores the environment, this trajectory gradually captures observations of objects, places, and events from different viewpoints and at different moments.

This accumulated experience contains information about the environment that may become useful later. During or after navigation, the robot may need to retrieve such information in order to support decision-making. For example, it may need to answer questions such as “Where is the exit?”, “Where can I buy a coffee?”, or “Which area did I visit last?”. The answers to these questions correspond to information previously observed along the navigation trajectory.

These queries are not abstract. They directly support navigation behavior: determining where to move next, verifying whether a location has already been visited, or recalling how to return to a previously observed place. In this sense, answering questions about past observations becomes a mechanism for accessing information stored in the robot’s navigation experience.

Answering such questions requires the system to refer back to the recorded trajectory, identify the relevant moment or several moments, and ground the response in visual evidence. This motivates the task of spatio-temporal video question answering over long egocentric trajectories, where a system must retrieve information from extended navigation experience while explicitly localizing the temporal intervals that support each answer. In this setting, the system receives a long egocentric video trajectory and a natural-language question, it must retrieve the supporting temporal moments and, when necessary, predict the answer. In this formulation, the navigation trajectory is stored once as persistent memory and can be queried repeatedly to retrieve evidence for different questions. This perspective complements vision-and-language navigation formulations that emphasize online action selection in interactive simulators~\citep{anderson2018matterport3d}.

Solving this problem requires more than instantaneous perception. Relevant evidence may appear only briefly, reoccur from different viewpoints, or become meaningful only when observations from distant points in time are combined. A system must therefore detect when relevant evidence appears, reason about spatial relationships between objects and places, understand semantic attributes and affordances, and aggregate information across long temporal horizons. Broader embodied benchmarks likewise stress grounding answers in accumulated egocentric experience~\citep{majumdar2024openeqa}.

Existing approaches face important limitations in this setting. Full-context vision--language models can achieve strong answer quality by processing the entire video sequence at once, but such approaches become computationally expensive and slow as video duration increases, even for recent long-context multimodal models such as Gemini~2.5 and Qwen3 ~\citep{kaduri2025s,liu2025survey,comanici2025gemini,bai2025qwen2, yang2025qwen3, gemini2025flash}. Moreover, repeatedly processing the full video for every query prevents the system from maintaining a persistent memory that can be reused efficiently. On the other hand, models designed for short video clips or compact segment descriptions often lose global context when applied to long trajectories \citep{lavila2023cvpr,fan2024videoagent_memory}.

These limitations motivate an alternative perspective: treating navigation experience as persistent structured memory that can be queried repeatedly. Embodied navigation research already emphasizes organizing observations into persistent spatial representations~\citep{tiddi2022knowledge}, including hierarchical object--zone priors for exploration~\citep{Zhang2021HOZ}, metric--semantic scene graphs for robotic mapping~\citep{armeni20193d,hughes2022hydra,Maggio2024Clio,3dgraphllm2025iccv,fross2025iccv}, and topological graphs for scalable place connectivity~\citep{garg2024robohop,chiang2024mobility,gsavln2025iclr}. These approaches demonstrate how long-horizon experience can be organized into structured representations. However, most of them are designed primarily to support planning and movement rather than retrospective question answering grounded in visual evidence over a completed trajectory.

VL-MemKnG combines a spatio-temporal knowledge graph with persistent segment-level contextual memory for temporally grounded question answering over long egocentric trajectories. The knowledge graph preserves structured relational information, while segment-level memory retains broader temporal and contextual continuity, enabling more reliable retrieval and aggregation of evidence distributed across distant temporal moments.

Language-guided navigation systems leverage vision--language models for semantic exploration---for example, frontier scoring for object-goal navigation~\citep{Yokoyama2024VLFM}---but typically reason over immediate observations and lack persistent, queryable memory of long trajectories for retrospective question answering. Retrieval-augmented pipelines combine captions, embeddings, and graph structure for scalable reasoning~\citep{wang2025navrag,edge2024graphrag}.

At the same time, recent memory-oriented approaches for long video understanding construct segment-level video memories using captioning and embedding indexing \citep{anwar2024remembr,fan2024videoagent_memory,kahatapitiya-etal-2025-language,you2024fdvs}. Such systems separate memory construction from query-time reasoning, enabling scalable retrieval over long videos. While effective for descriptive queries, these memories typically remain largely unstructured, which limits their ability to support explicit spatial reasoning and relational aggregation across distant temporal intervals.

In this work, we treat accumulated navigation experience as a persistent and queryable knowledge resource. We introduce VL-MemKnG, built on VL-KnG~\citep{almdfaa2025vlkng}, a hybrid memory architecture that combines segment-level retrieval with an explicit spatio-temporal knowledge graph encoding objects, attributes, and spatial relationships observed along the trajectory. Video segments act as temporal anchors for retrieval, while the knowledge graph enables relational reasoning and structured aggregation of evidence across time. The system supports two evidence formats: flat ranked frame lists for questions grounded in a single moment, and grouped frame lists for questions that require aggregating complementary evidence from multiple non-cooccurring temporal intervals.

A key design principle of the proposed system is that it is memory-centric. Computationally expensive operations—including video segmentation, caption generation, embedding computation, and graph construction—are performed offline only once during memory construction. At query time, the system retrieves and reasons over the resulting persistent memory instead of reprocessing the raw video. Graph retrieval and segment-level retrieval operate in parallel and their results are fused during reasoning. This separation enables efficient answering of multiple queries over long navigation trajectories.

To evaluate this setting, we use WalkieKnowledgeT+, an extension of the WalkieKnowledge benchmark~\citep{almdfaa2025vlkng} built on EgoWalk trajectories~\citep{Akhtyamov2025Egowalk}. The extension introduces temporally scattered and chronology-aware questions requiring multi-interval evidence grounding, following aggregation-style phenomena studied in grounded multi-hop VideoQA~\citep{chen2024gelm_multihop_egovqa}. These tasks test whether a system can retrieve and organize evidence distributed across non-contiguous points in the trajectory rather than relying on a single moment.

The resulting problem imposes four requirements on the system. First, answers must be grounded in explicit visual evidence rather than produced as unsupported text. Second, relevant information may be distributed across long temporal horizons and across multiple distinct moments. Third, the memory representation must remain persistent and efficiently queryable. Fourth, the reasoning process should remain interpretable, which is important in robotic settings where retrieved evidence may influence subsequent actions.

The goal of this work is therefore to develop and analyze a practical system for spatio-temporal question answering over long egocentric navigation trajectories that combines persistent structured memory with scalable retrieval and evidence-grounded reasoning.

Our contributions are threefold:
\begin{itemize}

\item We propose \textbf{VL-MemKnG}, a hybrid memory framework that combines a spatio-temporal knowledge graph with persistent segment-level contextual memory, enabling complementary retrieval of structured relational cues and long-horizon temporal context for temporally grounded question answering over long egocentric trajectories.
\item We introduce \textbf{WalkieKnowledgeT+}, an extended benchmark with temporally scattered and chronology-aware questions that require multi-interval evidence grounding and long-horizon temporal reasoning.
\item Extensive experiments demonstrate that VL-MemKnG improves Top-1 evidence retrieval and downstream QA performance on temporally distributed reasoning tasks while maintaining efficient query-time inference compared to graph-based and full-context vision--language baselines.

\end{itemize}

\section{Related Work}

This section reviews prior work related to reasoning over long egocentric navigation trajectories. We focus on approaches that construct persistent representations of navigation experience, enable retrieval over long visual histories, and support reasoning about spatial relationships and temporally distributed evidence.

Existing work relevant to this problem can be broadly grouped into four directions: structured navigation representations, graph-based environment models, retrieval-augmented memory frameworks, and long-video reasoning systems.

\subsection{Navigation and structured environment representations}

Embodied navigation research has long emphasized that purely reactive perception is insufficient in large, partially observable environments. When the target object is not visible, an agent must rely on persistent representations of past observations and prior knowledge about how objects and places are organized. A recurring idea in this line of work is that observations should be organized into structured environment representations rather than treated as isolated frames.

Early approaches therefore introduced semantic abstractions that guide exploration. \citet{Zhang2021HOZ}, for example, proposed the Hierarchical Object-to-Zone (HOZ) Graph, which organizes environments into a hierarchy of scenes, zones, and objects and performs exploration in a coarse-to-fine manner. Instead of directly searching for a target object, the system first identifies a relevant semantic zone and then uses it as a sub-goal for navigation.

More recent methods leverage vision--language models to guide exploration in previously unseen environments. \citet{Yokoyama2024VLFM} introduced Vision-Language Frontier Maps (VLFM), which combine classical frontier exploration with semantic relevance computed using vision--language similarity. By ranking candidate exploration frontiers based on their semantic similarity to the target description, the system enables zero-shot semantic navigation without task-specific training.

Vision-and-language navigation (VLN) more broadly integrates perception, language, and action for tasks such as instruction following and object search~\citep{anderson2018matterport3d}. Unlike most vision–language navigation settings that focus on online action selection, our work treats navigation experience as a recorded egocentric trajectory that can be queried retrospectively.

\subsection{Egocentric QA Benchmarks}

Several benchmarks have recently been introduced for question answering over egocentric observations and navigation experiences.

Benchmarks such as OpenEQA~\citep{majumdar2024openeqa} further highlight the importance of grounding language queries in accumulated egocentric experience. While these benchmarks evaluate reasoning about past observations, they typically focus on answer correctness rather than explicitly retrieving temporal evidence intervals that justify the answer. 

NaVQA~\citep{anwar2024remembr} focuses on navigation-oriented question answering in driving environments and provides 210 question-answer pairs across seven ego-centric driving sequences from CODa~\citep{zhang2024toward}. Similar to OpenEQA, the benchmark evaluates answer prediction rather than evidence localization and ranking.

\subsection{Semantic and graph-based environment representations}

Persistent semantic environment representations are central to many robotic perception systems. Among them, scene graphs provide a structured abstraction that represents environments through objects, attributes, and spatial relationships~\citep{armeni20193d}.

\citet{hughes2022hydra} introduced Hydra, a system that constructs a layered metric--semantic 3D scene graph from multimodal perception. The representation includes objects, places, and rooms connected through spatial relations, enabling persistent environment modeling across long deployments. Similarly, \citet{Maggio2024Clio} demonstrated how open-vocabulary vision--language models can guide which parts of a scene graph are semantically relevant for a task. Recent learning-based scene graph methods further integrate language models for online 3D scene understanding~\citep{3dgraphllm2025iccv,fross2025iccv}.

A complementary direction explores topological representations that scale to large environments without requiring metric reconstruction. \citet{garg2024robohop} proposed RoboHop, which represents a monocular navigation trajectory as a graph whose nodes correspond to semantically meaningful image segments and whose edges encode connectivity between observations. Queries can then retrieve relevant locations using vision--language similarity. Related graph-enhanced navigation approaches pursue complementary trade-offs~\citep{chiang2024mobility,gsavln2025iclr}.

These approaches illustrate the value of persistent environment structure for navigation. However, most of them focus on spatial understanding and planning rather than retrospective reasoning over egocentric video trajectories with temporally grounded evidence. Unlike many scene-graph mapping systems that rely on depth sensing and 3D reconstruction, our approach operates directly on monocular RGB trajectories.

VL-MemKnG bridges two families of representations: like scene graphs it stores objects with attributes and explicit spatial relations, and like topological graphs it scales to long monocular trajectories without requiring full 3D reconstruction.

\subsection{Retrieval-augmented and memory-based reasoning}

Long egocentric trajectories produce large volumes of observations that cannot be processed directly by models with limited context windows. As a result, recent work increasingly studies memory-centric reasoning frameworks that construct persistent memory representations and retrieve relevant evidence at query time.

Retrieval-augmented generation (RAG) conditions language-model reasoning on retrieved evidence rather than relying solely on parametric knowledge~\citep{lewis2020retrieval}. Graph-based extensions such as GraphRAG~\citep{edge2024graphrag} retrieve evidence from structured graph stores, enabling reasoning over relational knowledge. Retrieval-style pipelines also connect language goals to stored experience and intermediate representations~\citep{wang2025navrag}.

In embodied settings, several works construct persistent video memories. ReMEmbR~\citep{anwar2024remembr} stores navigation observations as captioned segments indexed by multimodal embeddings. At query time, relevant segments are retrieved and provided to a language model to generate answers. This separation between offline memory construction and online reasoning enables scalable question answering over long trajectories, but the stored information remains largely unstructured.

Prior VideoRAG and graph-based retrieval systems typically operate over clip-, caption-, or scene-level abstractions. In contrast, VL-MemKnG combines structured graph retrieval with persistent segment-level contextual memory for long-horizon temporally grounded reasoning.

\citet{almdfaa2025vlkng} addressed this limitation with VL-KnG, which organizes video observations into a spatio-temporal knowledge graph containing objects, attributes, and relations. Objects observed in different temporal segments are linked through spatiotemporal association, enabling persistent relational reasoning across the trajectory. Queries retrieve relevant subgraphs using GraphRAG-style pipelines and produce answers grounded in retrieved frames.

Our method, VL-MemKnG, extends this framework by combining graph-based relational memory with segment-level multimodal retrieval. This hybrid representation enables both relational reasoning over graph structure and contextual retrieval from video segments.

\subsection{Memory-augmented long-video reasoning}

Another line of research studies how systems can reason over long videos by constructing explicit intermediate memories rather than processing the full video context. Agentic systems such as VideoAgent~\citep{fan2024videoagent_memory} use large language models to control retrieval tools that access temporal and object memories.

Other approaches represent long videos primarily through language. LangRepo~\citep{kahatapitiya-etal-2025-language} constructs a repository of textual memory entries derived from video captions, enabling language-based retrieval and reasoning. Similarly, FDVS~\citep{you2024fdvs} generates hierarchical textual summaries that compress long videos into multi-scale narrative descriptions.

While such representations enable scalable reasoning over long videos, they typically rely on caption-based or narrative memory and do not explicitly maintain persistent object identities or spatial relations in a queryable graph structure. For navigation-oriented reasoning, where spatial relations and object continuity are important, structured relational memory becomes particularly valuable.

\subsection{Multi-hop reasoning and temporally scattered evidence}

A related challenge in video question answering concerns multi-hop reasoning across temporally distant observations. Some questions cannot be answered from a single moment and instead require aggregating evidence across multiple non-contiguous intervals.

\citet{chen2024gelm_multihop_egovqa} studied grounded multi-hop VideoQA and introduced mechanisms for retrieving multiple temporal evidence spans. Their work highlights the importance of explicitly retrieving temporally scattered evidence.

Motivated by this observation, our formulation distinguishes between two evidence structures: flat rankings for single-moment questions and grouped evidence when answers require aggregating information from multiple temporal intervals. VL-MemKnG supports both evidence formats by combining segment-level retrieval with graph-based relational reasoning.

Across these research directions, several common design principles emerge:
(i) persistent structured representations of the environment,
(ii) offline memory construction with efficient query-time retrieval,
(iii) hybrid reasoning that combines relational structure with multimodal retrieval, and
(iv) explicit grounding of answers in temporally localized evidence.

VL-MemKnG builds on these principles by integrating segment-level multimodal memory with a persistent spatio-temporal knowledge graph and applying retrieval-augmented reasoning to answer navigation-oriented questions over long egocentric trajectories.

\section{Materials and Methods}
\label{sec:materials_methods}

\subsection{Task Definition}
\label{subsec:problem_formulation}

We study question answering over long egocentric navigation trajectories with explicit temporal evidence grounding. A navigation episode is represented as a video $\mathcal{V} = \{I_t\}_{t=1}^{T}$,
where $I_t$ denotes the frame corresponding to timestep $t$.

To enable efficient reasoning over long videos, we operate on a sparsely sampled subset of frames $\tilde{\mathcal{V}} \subset \mathcal{V}$, where
$\tilde{\mathcal{V}} = \{I_t\}_{t\in\mathcal{T}}$ and $\mathcal{T}\subseteq\{1,\dots,T\}$.

For each question $q$, let
$
\mathcal{F}(q)
=
\{[t^s_i,t^e_i]\}_{i=1}^{N_q}
$
denote the ground-truth temporal evidence intervals associated with the information required to answer the question.

For questions requiring multiple distinct supporting moments, the ground-truth evidence may additionally be represented as grouped interval evidence
$$
\mathcal{F}(q)
=
\{
\mathcal{F}^{(1)}_q,
\dots,
\mathcal{F}^{(M)}_q
\},
$$
where each group
$
\mathcal{F}^{(m)}_q
=
\{[t^s_i,t^e_i]\}_{i=1}^{N_m}
$
contains intervals corresponding to one component of the required evidence.

Given a natural language question $q$, the goal is to predict relevance-ranked supporting evidence frames $\mathcal{F}^*(q)$, grounded in sampled frame indices
    $
    \{\, t \mid I_t \in \tilde{\mathcal{V}} \,\}
    \subseteq
    \{1,\dots,T\}
    $,
that aligns with the ground-truth evidence $\mathcal{F}(q)$, and to provide an answer $a^* \in C(q)$ conditioned on the predicted evidence, where $C(q)$ denotes the candidate answer set associated with question $q$.

For questions grounded in a single temporal moment, $\mathcal{F}^*(q)$
is represented as a flat ranked list of frame indices $[t_1,\dots,t_K]$, and $t_i \in \{1,\dots,T\}$.

For questions requiring evidence from multiple non-cooccurring temporal intervals,
$$
\mathcal{F}^*(q) = \{\mathcal{F}_q^{*(1)}, \dots, \mathcal{F}_q^{*(M)}\}.
$$
where each group $\mathcal{F}_q^{*(m)}$ contains frames supporting one component of the answer, ranked according to their estimated relevance within that temporal group.

Relevant evidence may be distributed across temporally distant intervals, requiring retrieval, grounding, and aggregation of complementary information from multiple parts of the trajectory.

The task therefore requires jointly performing temporal evidence localization and answer prediction over both segment-level memory and a spatio-temporal knowledge graph.

Formally, VL-MemKnG performs two stages. First, a retrieval stage
\begin{equation}
\phi_{\mathrm{retr}} :
(q, \tilde{\mathcal{V}}, \mathcal{S}, \mathcal{G},)
\rightarrow
(\mathcal{S}_q, \mathcal{G}_q),
\end{equation}
retrieves a set of question-relevant memory segments
$\mathcal{S}_q \subseteq \mathcal{S}$ and a set of question-relevant evidence subgraphs $\mathcal{G}_q \subseteq \mathcal{G}$.

Second, a reasoning stage
\begin{equation}
\phi_{\mathrm{ans}} :
(q, \tilde{\mathcal{V}}, \mathcal{S}_q, \mathcal{G}_q)
\rightarrow
(\mathcal{F}^*(q), a^*),
\end{equation}
uses a large language model to jointly perform temporal evidence grounding and answer prediction over the retrieved graph and segment-level evidence.

Here, $\mathcal{S}$ denotes the segment-level memory, and $\mathcal{G}$ denotes the spatio-temporal knowledge graph.

\subsection{VL-MemKnG Framework}
\label{sec:method}

We present \emph{VL-MemKnG}, a hybrid memory framework for navigation-oriented question answering over long egocentric trajectories with explicit temporal evidence grounding. The method combines two complementary memory representations: (i) a segment-level memory that provides contextual temporal anchors through captions and multimodal embeddings, and (ii) a spatio-temporal knowledge graph that explicitly represents objects, attributes, and spatial relations observed throughout the trajectory.

VL-MemKnG follows a memory-centric design. The raw video is processed only once during an offline phase where persistent memory structures are constructed. During the online phase, questions are answered by retrieving and reasoning over these precomputed memories rather than repeatedly processing the original video stream. This design enables efficient retrieval and reasoning over long trajectories and supports repeated querying.

VL-MemKnG builds directly on the VL-KnG framework~\citep{almdfaa2025vlkng}, inheriting its caption-derived spatio-temporal knowledge graph, cross-chunk identity consistency, and GraphRAG-style graph retrieval pipeline. From long-video reasoning systems such as VideoAgent~\citep{fan2024videoagent_memory}, it adopts the design pattern of segment-level memory indexed by captions and multimodal embeddings. The main methodological contribution of VL-MemKnG is to integrate these two memory representations into a unified architecture where graph-structured relational retrieval and segment-level contextual retrieval operate in parallel and are fused during reasoning.

\subsection{Method Overview}

VL-MemKnG operates in two phases: an offline memory construction phase and an online question answering phase.

\textbf{Offline phase.}
The navigation trajectory is partitioned into temporal segments. Captions and multimodal embeddings are generated for each segment. Together, these representations form a persistent segment-level memory $\mathcal{S}$ and are further used to construct a spatio-temporal knowledge graph $\mathcal{G}$.

\textbf{Online phase.}
Given a natural-language question, the system retrieves relevant evidence subgraphs and memory segments from both memory structures in parallel. The retrieved evidence is anchored to temporal intervals in the trajectory, which are subsequently mapped to the sparsely sampled frames used for reasoning. The resulting frame-level evidence is then processed by a reasoning module that ranks supporting evidence and predicts the final answer.

\begin{figure}[h!]
\begin{center}
\includegraphics[width=\textwidth]{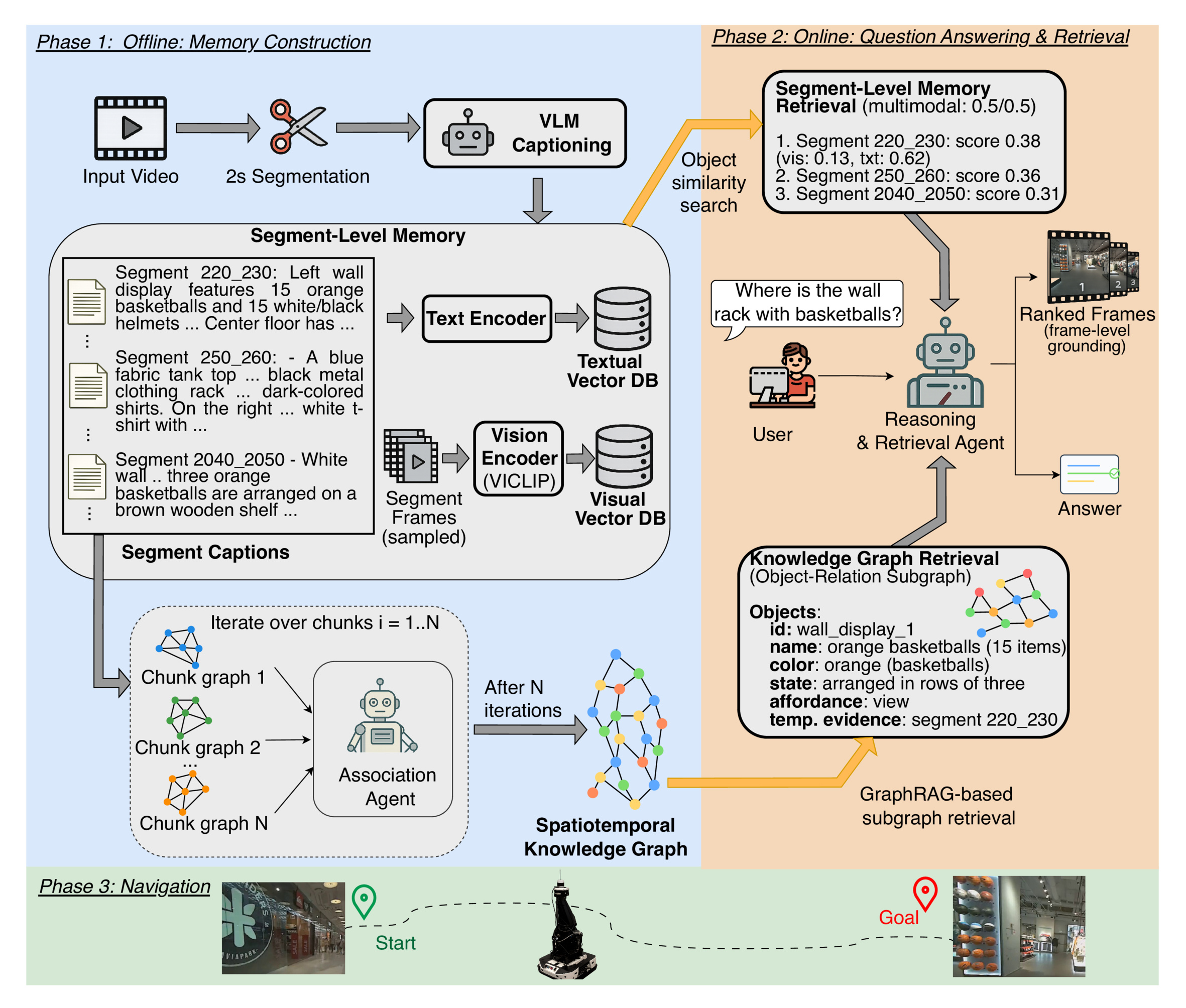}
\end{center}
\caption{Overview of VL-MemKnG. The offline phase constructs segment-level memory and a spatio-temporal knowledge graph from the navigation trajectory. During the online phase, both memories are retrieved in parallel and fused to produce answers and temporally grounded evidence.}
\label{fig:vlmemkng_pipeline}
\end{figure}

At a high level, the pipeline consists of four stages (Figure~\ref{fig:vlmemkng_pipeline}): (1) video segmentation and caption generation, (2) construction of segment-level multimodal memory, (3) construction of a spatio-temporal knowledge graph, and (4) hybrid retrieval and reasoning for temporal evidence grounding and answer prediction.

\subsection{Video Segmentation and Caption Generation}
\label{sec:segmentation_captioning}

During the offline phase, the navigation trajectory is partitioned into contiguous fixed-length temporal segments. Each segment is assigned a unique identifier encoding its frame range. In our implementation, each segment corresponds to approximately two seconds of trajectory time in the 5 fps videos.

For each segment, a small set of representative frames is uniformly sampled and passed to a vision-language model to generate a caption describing the visual scene. Unlike narrative captions commonly used for video summarization, these captions are designed as structured intermediate representations that emphasize visually grounded objects, attributes, readable scene text, and spatial relationships.

The segment-level memory construction follows the long-video retrieval framework introduced in VideoAgent~\citep{fan2024videoagent_memory}, where videos are represented through segment captions and multimodal embeddings. Unlike the original VideoAgent pipeline, which uses LaViLa~\citep{lavila2023cvpr} for compact narrative summarization, VL-MemKnG generates captions using a vision--language model with constrained object-centric prompts that encourage explicit descriptions of objects, attributes, and spatial configurations (Prompt~\ref{box:caption_prompt}). These captions serve both as retrieval anchors and as input for knowledge-graph construction.

\newcounter{prompt}
\definecolor{promptblue}{RGB}{0,51,102}
\definecolor{promptteal}{RGB}{0,109,109}
\definecolor{promptalert}{RGB}{200,40,40}
\newcommand{\pcritical}[1]{\textbf{\textcolor{promptalert}{#1}}}
\refstepcounter{prompt}
\label{box:caption_prompt}
\fbox{%
\begin{minipage}{0.94\linewidth}
\small
{\color{promptblue}\textbf{Prompt Template}}
\vspace{0.4em}

{\color{promptblue}\textbf{Task:}} 
For each segment, produce a compact object-centric caption that serves as a temporal anchor for retrieval and knowledge-graph construction, rather than a narrative summary.
\vspace{0.4em}

{\color{promptteal}\textbf{Critical rules}}

\vspace{-0.4em}
\begin{list}{\textcolor{promptteal}{--}}{
  \setlength{\leftmargin}{1.1em}
  \setlength{\labelwidth}{0.8em}
  \setlength{\labelsep}{0.3em}
  \setlength{\itemsep}{0pt}
  \setlength{\parsep}{0pt}
  \setlength{\topsep}{0.2em}
}
\item \pcritical{No} storytelling or camera/perspective commentary.
\item \pcritical{Describe repeated objects separately:} each instance must differ by at least one attribute.
\item For every visible object, include relevant attributes when available, such as \texttt{category}, \texttt{color}, \texttt{material}, \texttt{size}, \texttt{state}, \texttt{position}, \texttt{area}, and \texttt{affordance}.
\item \pcritical{Preserve all readable text exactly}.
\end{list}
\vspace{0.2em}
{\color{promptteal}\textbf{Spatial relation vocabulary.}}
{\texttt{on, on\_top\_of, under, next\_to, between, in\_front\_of, behind, near, far\_from, touching, separate\_from, left\_of, right\_of, above, below, inside, outside, surrounding, adjacent\_to, against}}

\end{minipage}%
}

{\small
\textbf{Prompt \theprompt.}
Prompt Template used for object-centric segment caption generation.
}

\subsection{Segment-Level Memory}
\label{sec:segment_memory}

Segment-level memory stores captioned temporal segments together with their multimodal representations for retrieval during query processing.

Each segment is represented by:
\begin{itemize}
\item a segment identifier $s_i = [t_i^{start}, t_i^{end})$, encoding the frame range covered by the segment,
\item a caption $c_i$ describing the scene,
\item a text embedding $\mathbf{e}^{\mathrm{text}}_i$ computed from the caption, and
\item a visual embedding $\mathbf{e}^{\mathrm{vis}}_i$ extracted from a short video clip.
\end{itemize}

Formally, the segment memory is defined as $
\mathcal{S} = \{(s_i, c_i, \mathbf{e}^{\mathrm{text}}_i, \mathbf{e}^{\mathrm{vis}}_i)\}_{i=1}^{N}.$

Textual embeddings encode semantic representations of segment captions for retrieval based on similarity to natural-language queries, while visual embeddings encode visual representations of trajectory segments for retrieval based on visual similarity.

Segment textual embeddings are computed using \texttt{text-embedding-3-large}, while visual embeddings are extracted using ViCLIP~\citep{viclip2023} from sparsely sampled frames within each segment.

\subsection{Spatio-Temporal Knowledge Graph Construction}
\label{sec:kg_construction}

VL-MemKnG constructs a spatio-temporal knowledge graph from the generated segment captions to support structured reasoning over objects, attributes, and spatial relationships.

Formally, the knowledge graph is represented as $\mathcal{G} = (\mathcal{O}, \mathcal{R})$, where $\mathcal{O}$ denotes the set of object nodes with semantic attributes, and $\mathcal{R}$ denotes the set of spatial relations between objects.

Following the VL-KnG framework~\citep{almdfaa2025vlkng}, captions are processed in fixed-size chunks and provided to a large language model to extract objects, semantic attributes, and spatial relations. Each chunk yields a local graph representation of the observed scene. VL-MemKnG further adopts the spatio-temporal object association (STOA) mechanism introduced in VL-KnG to associate repeated observations of the same object across different chunks within a shared global graph.

\subsection{Hybrid Retrieval}
\label{sec:hybrid_retrieval}

\subsubsection{Segment-level Memory Retrieval}

The input question is transformed into an object-centric query emphasizing visually grounded entities relevant for retrieval. For example, the question \emph{What is to the left of the children's shopping carts with toy cars in front?''} is reduced to \emph{children's shopping carts with toy cars in front''}. Additional analysis of query formulation choices is provided in Section \ref{subsec:ablation_query_embed}.

The resulting query is embedded into textual and visual representation spaces and matched against the segment-level memory using similarity search.

Textual and visual similarity scores are combined using weighted linear fusion,
$$s(i) = w_{vis} \cdot s_{vis}(i) + w_{text} \cdot s_{text}(i),$$
where $s_{\mathrm{vis}}(i)$ and $s_{\mathrm{text}}(i)$ denote the visual and textual similarity scores for segment $(i)$. 

In the current implementation, equal weights are used for textual and visual retrieval: $ w_\text{vis} = 0.5$,  $w_\text{text} = 0.5$. Additional analysis is provided in Section \ref{subsec:ablation_fusion}.

Segments are ranked according to the fused similarity score, producing a set of retrieved segments
$\mathcal{S}_q \subseteq \mathcal{S}$.

\subsubsection{Knowledge graph retrieval}

Following the retrieval procedure introduced in VL-KnG~\citep{almdfaa2025vlkng}, natural-language questions are adapted to align with object names and semantic descriptions present in the knowledge graph while preserving the meaning of the original question. The adapted query is then used to retrieve relevant segment-anchored subgraphs containing candidate objects, semantic attributes, and spatial relations. The retrieved subgraphs form a set of evidence subgraphs $\mathcal{G}_q \subseteq \mathcal{G}$.

\subsection{Frame-Level Evidence Grounding}
\label{sec:segment_to_frame_grounding}

Both retrieval streams return segment-anchored evidence, whereas the benchmark requires evidence to be reported as frame indices.

Each segment identifier corresponds to a temporal interval in the trajectory. To obtain frame-level evidence, the retrieved segments are therefore mapped to the subset of sparsely sampled frames $\tilde{\mathcal{V}}$ introduced in Section~\ref{subsec:problem_formulation}. The sampling protocol used to construct this frame set is described in Section~\ref{subsec:frame_decimation}.

This procedure converts segment-level evidence into a frame-level representation while preserving temporal alignment with the original trajectory. Importantly, frame-level grounding serves not only to map segment-level evidence to frame indices. Because evaluation is performed at the frame level, sparsely sampled frames reduce redundancy between temporally adjacent observations and avoid dense continuous evidence spans, producing a more discriminative setting for evidence localization and frame ranking over long trajectories.

\subsection{Reasoning and Evidence Ranking}
\label{sec:reasoning}

After retrieval and frame-level grounding, VL-MemKnG performs joint reasoning over the combined evidence context.

The reasoning stage jointly processes two complementary evidence sources: segment-level evidence retrieved from memory, consisting of captions and their associated grounded frames; and graph evidence retrieved from the knowledge graph, including objects, semantic attributes, spatial relations, and associated grounded frames.

The language model analyzes the retrieved evidence to rank supporting frames and, when applicable, predict the final answer.

Depending on the temporal structure of the question, the system produces either a \textbf{flat ranking}, represented as a single ranked list of frames for questions whose evidence is concentrated within one temporal region, or a \textbf{grouped ranking}, represented as multiple temporally distinct frame groups for questions whose answer depends on evidence aggregated across multiple non-cooccurring temporal intervals.

This unified reasoning framework enables VL-MemKnG to support both single-moment evidence localization and multi-interval temporal reasoning over long navigation trajectories.

\section{WalkieKnowledgeT+ Benchmark}
\label{sec:benchmark}

We evaluate the proposed framework on \emph{WalkieKnowledgeT+}. The benchmark contains eight long egocentric navigation trajectories covering both indoor and outdoor environments and includes 262 natural-language questions annotated with temporally grounded supporting evidence.

WalkieKnowledgeT+ follows the evaluation protocol of WalkieKnowledge to ensure direct comparability with prior results while extending the benchmark with additional temporal reasoning tasks.

\begin{figure}[h!]
\begin{center}
\includegraphics[width=\textwidth]{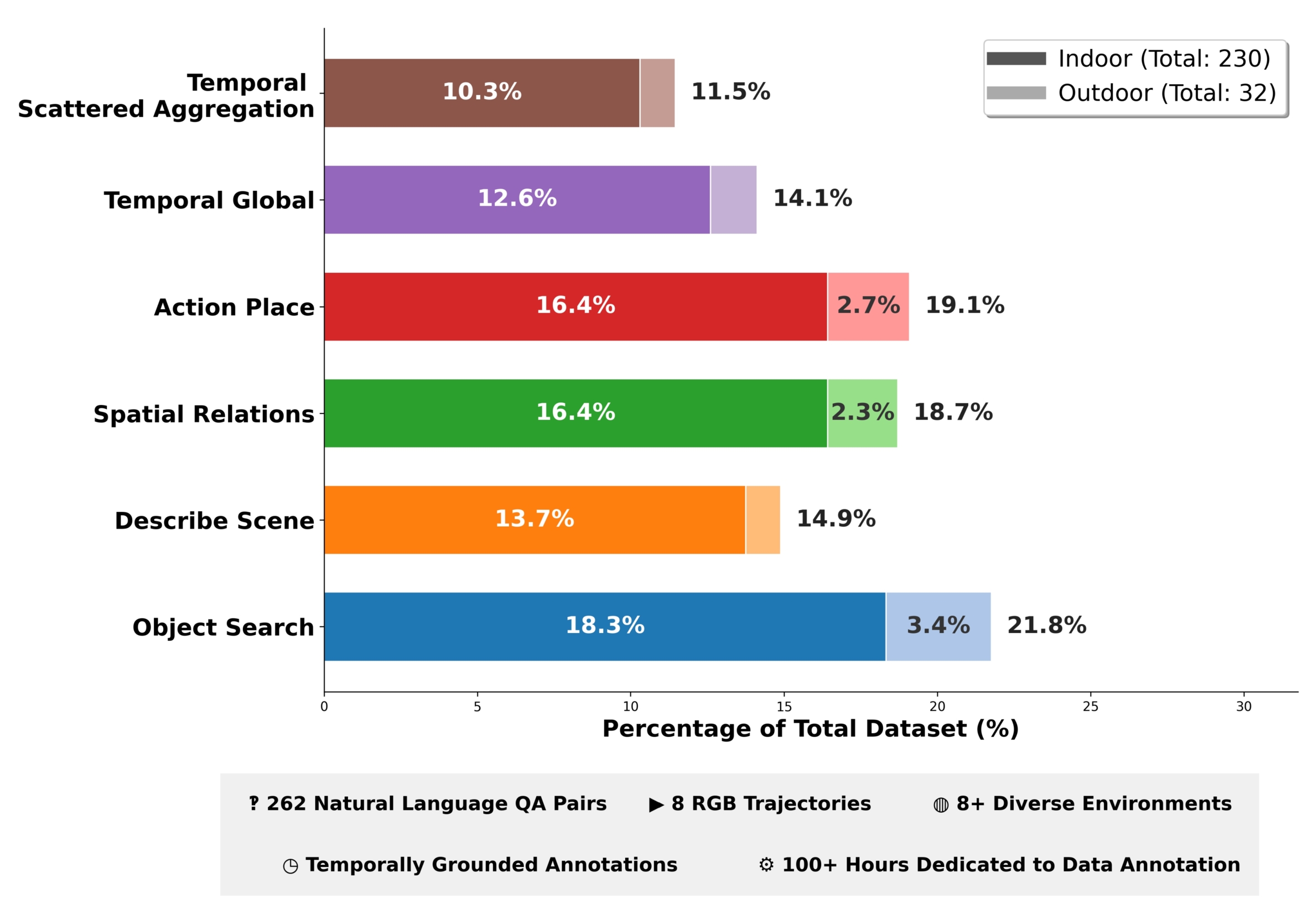}
\end{center}
\caption{Distribution of \emph{WalkieKnowledgeT+} questions by category and environment. Each bar shows the share of the 262 question--answer pairs in a category, split between indoor and outdoor trajectories.}
\label{fig:wk_tplus_stats}
\end{figure}

Figure~\ref{fig:wk_tplus_stats} shows the distribution of question types across the dataset.

\subsection{Question Types}
\label{subsec:question_types}

In addition to standard navigation-oriented question types, including \emph{object search}, \emph{scene description}, \emph{spatial relations}, and \emph{action--place} association, WalkieKnowledgeT+ introduces two temporal reasoning categories: \emph{temporal-global} questions and \emph{temporally scattered aggregation} questions.

\emph{Temporal-global} questions require reasoning about chronology over the trajectory as a whole, such as determining which area was reached earlier or later. \emph{Temporally scattered aggregation} questions require aggregating evidence from multiple non-cooccurring moments, for example to identify all places satisfying a shared visual or semantic condition.

The benchmark supports two evidence formats. In the \textbf{flat} setting, supporting evidence is represented as a single ranked list of frame indices ordered by relevance to the query. In the \textbf{grouped} setting, evidence is represented as multiple temporally distinct frame groups, where each group corresponds to a separate supporting moment required for answering the question and frames within each group are ranked by relevance.

\begin{figure}[h!]
\begin{center}
\includegraphics[width=\textwidth]{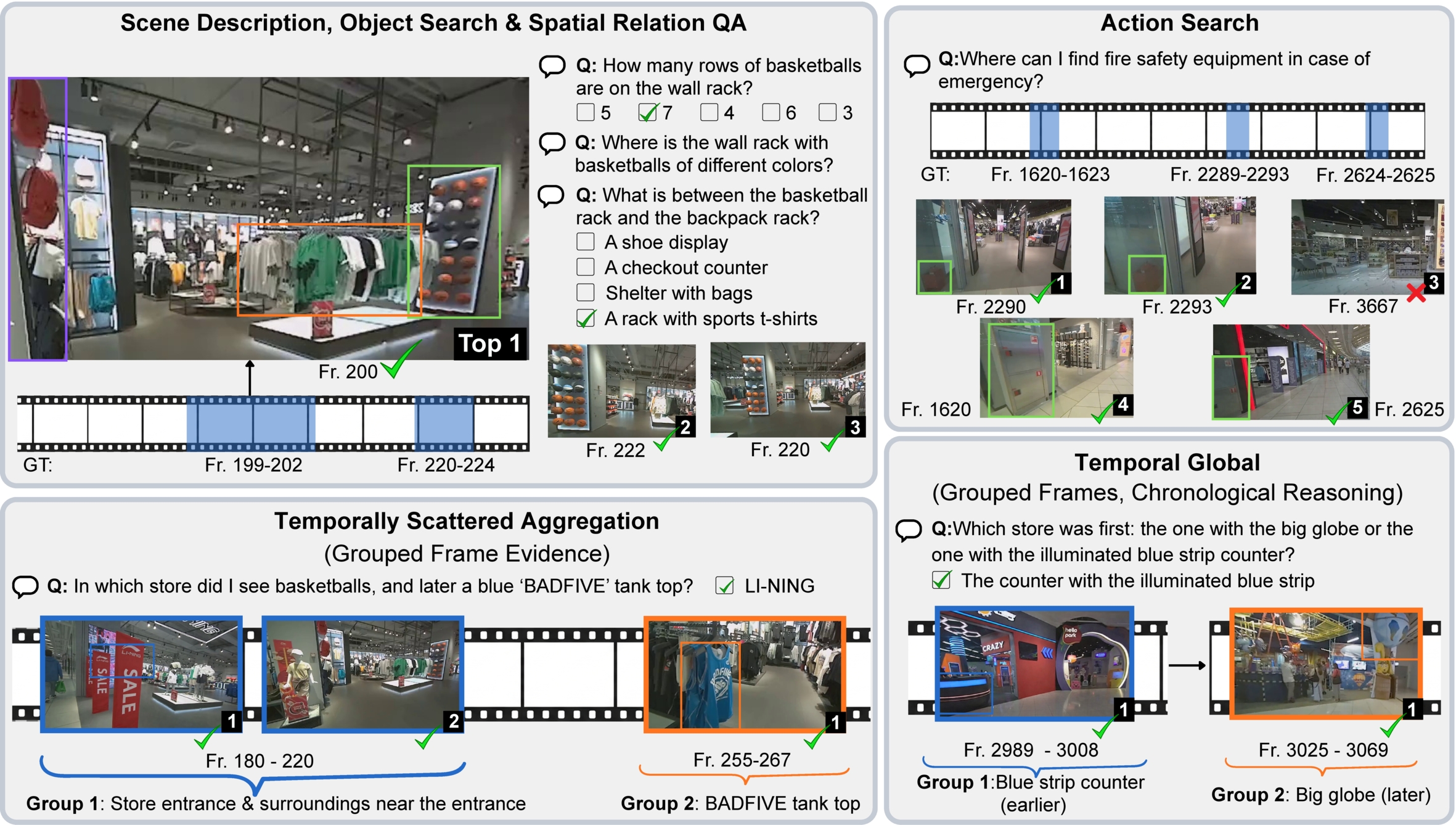}
\end{center}
\caption{Representative predictions across question categories and temporal evidence annotations from WalkieKnowledgeT+. Examples include scene description and spatial-relation reasoning, object and action--place localization, and temporal-global and temporally scattered aggregation questions, where evidence is organized across multiple non-cooccurring moments.}
\label{fig:qa_examples}
\end{figure}

Figure~\ref{fig:qa_examples} presents representative predictions across different question categories together with ground-truth temporal evidence annotations represented as flat or grouped frame intervals. For \emph{scene description} and \emph{spatial-relation} questions, VL-MemKnG retrieves frames containing the relevant objects and spatial configurations required for answering the question. For \emph{object search} and \emph{action--place} questions, the retrieved evidence localizes the target object or activity-associated location. For temporal-global and temporally scattered aggregation questions, the system organizes evidence into multiple temporally distinct groups corresponding to different supporting moments along the trajectory.

\subsection{Frame Decimation}
\label{subsec:frame_decimation}

Long egocentric trajectories contain substantial temporal redundancy. Following \citet{almdfaa2025vlkng}, we apply uniform frame decimation for downstream visual processing by selecting every 60th frame from videos sampled at 5 fps, enabling tractable frame-level retrieval and ranking during evaluation.

Uniform decimation may omit all sampled frames from a ground-truth evidence interval. To ensure fair evaluation, if no sampled frame falls within a question's annotated evidence interval, we additionally retain at least one frame from that interval. This guarantees that all questions remain answerable while keeping the number of processed frames low.

\subsection{Evaluation Protocol}
\label{subsec:evaluation_protocol}

Models are evaluated using both answer prediction and evidence retrieval metrics. \textbf{Answer Accuracy} is reported for multiple-choice questions and measures whether the system selects the correct answer option.

Evidence retrieval is evaluated by comparing predicted frame rankings with annotated ground-truth evidence intervals. Following \citet{almdfaa2025vlkng}, we report \textbf{Retrieval Accuracy@k}, which measures whether relevant evidence appears among the top-$k$ retrieved frames, together with \textbf{Precision@k}, \textbf{Recall@k}, and \textbf{MRR@k}.

For \emph{temporal-global} and \emph{temporally scattered aggregation} questions, evidence may span several non-cooccurring temporal moments. In these cases, both annotations and predictions are represented as temporally grouped frame evidence, where each group corresponds to a distinct supporting moment along the trajectory. Each annotated group is matched to the predicted group with the largest frame overlap, after which retrieval metrics are computed over the matched groups. Scores are first averaged across groups within each question and then across questions.

This evaluation protocol measures not only evidence retrieval accuracy, but also the ability to organize evidence distributed across multiple temporal intervals.

\section{Results}
\label{sec:results}

\subsection{Overall Performance}
\label{subsec:overall_results}

We compare VL-MemKnG against the graph-based VL-KnG variants~\citep{almdfaa2025vlkng}, RoboHop~\citep{garg2024robohop}, and full-context vision--language models including Qwen 3.5+, Gemini 2.5 Flash, and Gemini 2.5 Pro~\citep{yang2025qwen3, gemini2025flash}. All VL-MemKnG experiments use balanced multimodal fusion weights ($w_{\mathrm{vis}} = w_{\mathrm{text}} = 0.5$). RoboHop is evaluated only on the original WalkieKnowledge benchmark because it does not natively support grouped temporal evidence retrieval.

The main comparison on WalkieKnowledgeT+ is summarized in Table~\ref{tab:overall_performance_wktplus}. Compared with the strongest graph-only VL-KnG variant, VL-MemKnG improves Retrieval Accuracy@1 from 58.33\% to 66.83\%, Recall@1 from 34.50\% to 40.55\%, and MRR@1 from 0.583 to 0.668. Notably, VL-MemKnG achieves the strongest overall Top-1 retrieval performance across Retrieval Accuracy@1, Recall@1, Precision@1, and MRR@1 even compared with frontier full-context vision--language models.

\begin{table}[htbp]
\centering
\scriptsize
\setlength{\tabcolsep}{2pt}
\caption[Overall Performance, WalkieKnowledgeT+]{%
Overall performance on the WalkieKnowledgeT+ benchmark.
All metrics are reported as percentages (\%), except for Mean Reciprocal Rank (MRR). Higher is better ($\uparrow$). The top three results per row are highlighted by color:
\protect\colorbox{best}{1st},
\protect\colorbox{second}{2nd}, and
\protect\colorbox{third}{3rd}.
}
\label{tab:overall_performance_wktplus}
\begin{tabular}{lcccccc}
\toprule
\textbf{Metric} &
\multicolumn{1}{c}{\textbf{VL-MemKnG}} &
\multicolumn{2}{c}{\textbf{VL-KnG}} &
\textbf{Qwen 3.5+} &
\multicolumn{2}{c}{\textbf{Gemini 2.5}} \\
\cmidrule(lr){3-4}
\cmidrule(lr){6-7}
& \textbf{(Ours)}
& \textbf{GER-L} & \textbf{GER-G}
& & \textbf{Flash} & \textbf{Pro} \\
\midrule
\multicolumn{7}{l}{\textit{Retrieval Perf. (\%)}} \\
\quad Retrieval Acc.@1$\uparrow$
& \best{66.83} & 55.00 & 58.33 & \third{63.37} & 59.01 & \second{66.03} \\
\quad Retrieval Acc.@3$\uparrow$
& 72.30 & 70.90 & 70.90 & \second{78.59} & \third{77.60} & \best{85.06} \\
\quad Retrieval Acc.@5$\uparrow$
& 75.35 & 74.94 & 73.01 & \second{81.67} & \third{78.78} & \best{86.35} \\
\quad Recall@1$\uparrow$
& \best{40.55} & 32.22 & 34.50 & \third{39.69} & 37.22 & \second{40.50} \\
\quad Recall@3$\uparrow$
& 57.67 & 57.75 & \third{58.30} & \best{60.93} & 53.21 & \second{59.01} \\
\quad Recall@5$\uparrow$
& 60.82 & \second{62.50} & \third{61.60} & \best{65.05} & 54.53 & 60.68 \\
\quad Precision@1$\uparrow$
& \best{66.83} & 55.00 & 58.34 & \third{63.37} & 59.01 & \second{66.02} \\
\quad Precision@3$\uparrow$
& 36.90 & \second{38.23} & \best{38.66} & \third{38.07} & 31.42 & 35.29 \\
\quad Precision@5$\uparrow$
& 24.49 & \second{25.25} & \third{25.11} & \best{25.36} & 19.48 & 22.30 \\
\midrule
\multicolumn{7}{l}{\textit{Ranking Quality}} \\
\quad MRR@1$\uparrow$
& \best{0.668} & 0.550 & 0.583 & \third{0.634} & 0.590 & \second{0.660} \\
\quad MRR@3$\uparrow$
& \third{0.694} & 0.624 & 0.640 & \second{0.705} & 0.673 & \best{0.749} \\
\quad MRR@5$\uparrow$
& \third{0.701} & 0.634 & 0.645 & \second{0.712} & 0.676 & \best{0.752} \\
\midrule
\multicolumn{7}{l}{\textit{Generation Quality (\%)}} \\
\quad Answer Acc.
& 65.81 & 50.33 & 49.02 & \best{70.59} & \third{67.32} & \second{67.97} \\
\bottomrule
\end{tabular}
\end{table}

These improvements suggest that augmenting the spatio-temporal knowledge graph with segment-level memory improves evidence ranking quality, particularly at the highest retrieval positions. While full-context vision--language models remain competitive at larger values of $k$ and in answer prediction, VL-MemKnG more consistently retrieves relevant evidence among the highest-ranked frames without requiring repeated full-video processing at query time.

Table~\ref{tab:overall_performance} reports the same comparison on the original WalkieKnowledge benchmark. The overall trend remains consistent but is less pronounced than on WalkieKnowledgeT+, suggesting that the benefit of hybrid memory becomes more apparent on temporally structured reasoning tasks.

\begin{table}[htbp]
\centering
\scriptsize
\setlength{\tabcolsep}{2pt}
\caption[Overall Performance, WalkieKnowledge.]{
Overall performance on the WalkieKnowledge benchmark.
All metrics are reported as percentages (\%), except for Mean Reciprocal Rank (MRR). Higher is better ($\uparrow$).
The top three results per row are highlighted by color:
\protect\colorbox{best}{1st},
\protect\colorbox{second}{2nd}, and
\protect\colorbox{third}{3rd}.
}
\label{tab:overall_performance}
\begin{tabular}{lcccccccc}
\toprule
\textbf{Metric} &
\multicolumn{1}{c}{\textbf{VL-MemKnG}} &
\multicolumn{3}{c}{\textbf{VL-KnG}} &
\textbf{RoboHop} &
\textbf{Qwen 3.5+} &
\multicolumn{2}{c}{\textbf{Gemini 2.5}} \\
\cmidrule(lr){3-5}
\cmidrule(lr){8-9}
& \textbf{(Ours)}
& \textbf{GR}
& \textbf{GER-L}
& \textbf{GER-G}
& & & \textbf{Flash} & \textbf{Pro} \\
\midrule
\multicolumn{9}{l}{\textit{Retrieval Perf. (\%)}} \\
\quad Retrieval Acc.@1$\uparrow$
& \second{67.69} & 53.16 & 61.66 & 65.80 & 34.72 & \third{66.32} & 61.66 & \best{68.91} \\
\quad Retrieval Acc.@3$\uparrow$
& 73.85 & 62.11 & 80.31 & 81.35 & 54.40 & \second{83.42} & \third{82.90} & \best{88.08} \\
\quad Retrieval Acc.@5$\uparrow$
& 77.95 & 64.21 & \third{85.49} & 83.94 & 62.69 & \second{87.56} & 83.94 & \best{89.12} \\
\quad Recall@1$\uparrow$
& \third{40.05} & 28.28 & 35.28 & 38.54 & 19.28 & \best{41.15} & 37.76 & \second{40.94} \\
\quad Recall@3$\uparrow$
& 59.41 & 49.11 & \second{65.39} & \best{67.53} & 37.50 & \third{64.34} & 53.61 & 56.97 \\
\quad Recall@5$\uparrow$
& 63.50 & 52.32 & \second{71.11} & \best{71.75} & 47.40 & \third{69.78} & 54.72 & 58.13 \\
\quad Precision@1$\uparrow$
& \second{67.69} & 52.63 & 61.66 & 65.81 & 35.75 & \third{66.32} & 61.66 & \best{68.91} \\
\quad Precision@3$\uparrow$
& 39.14 & 34.91 & \second{44.21} & \best{45.60} & 24.35 & \third{40.59} & 32.12 & 34.54 \\
\quad Precision@5$\uparrow$
& 26.46 & 22.52 & \second{29.22} & \best{29.84} & 18.55 & \third{27.67} & 19.69 & 21.56 \\
\midrule
\multicolumn{9}{l}{\textit{Ranking Quality}} \\
\quad MRR@1$\uparrow$
& \second{0.677} & 0.526 & 0.617 & 0.658 & 0.347 & \third{0.663} & 0.617 & \best{0.689} \\
\quad MRR@3$\uparrow$
& 0.706 & 0.568 & 0.704 & \third{0.729} & 0.434 & \second{0.744} & 0.712 & \best{0.778} \\
\quad MRR@5$\uparrow$
& 0.715 & 0.572 & 0.716 & \third{0.735} & 0.454 & \second{0.754} & 0.714 & \best{0.781} \\
\midrule
\multicolumn{9}{l}{\textit{Generation Quality (\%)}} \\
\quad Answer Acc.
& \third{60.23} & 50.00 & 51.16 & 52.33 & 26.74 & \best{66.28} & \best{66.28} & \second{61.63} \\
\bottomrule
\end{tabular}
\end{table}

\subsection{Per-Category Analysis}
\label{subsec:per_category_analysis}

Table~\ref{tab:wk_tplus_per_category} reports the breakdown by question category on WalkieKnowledgeT+.

The largest improvements of VL-MemKnG over VL-KnG appear on \emph{temporal-global} and \emph{temporally scattered aggregation} questions, where evidence must be retrieved and organized across multiple non-cooccurring temporal moments. On temporal-global questions, VL-MemKnG achieves the strongest MRR@1 and MRR@3 among all compared methods. On temporally scattered aggregation questions, it substantially improves over graph-only VL-KnG variants and achieves the strongest answer accuracy.

\begin{table}[htbp]
\centering
\caption[WalkieKnowledgeT+: per-category]{Per-category evaluation on \emph{WalkieKnowledgeT+}, including temporal-global and temporal scattered aggregation. Recall@$k$ and answer accuracy are percentages; MRR@$k$ is not. $\uparrow$: higher is better. Per-column top three are shaded (\protect\colorbox{best}{1st} / \protect\colorbox{second}{2nd} / \protect\colorbox{third}{3rd}).}
\label{tab:wk_tplus_per_category}
\scriptsize
\setlength{\tabcolsep}{1.2pt}
\renewcommand{\arraystretch}{1.05}
\resizebox{\textwidth}{!}{%
\begin{tabular}{l*{22}{c}}
\toprule
\multirow{3}{*}{\textbf{Method}} &
\multicolumn{4}{c}{\textbf{Scene Desc.}} &
\multicolumn{4}{c}{\textbf{Spatial Rel.}} &
\multicolumn{3}{c}{\textbf{Obj.\ Search}} &
\multicolumn{3}{c}{\textbf{Action-Place}} &
\multicolumn{4}{c}{\textbf{Temp.\ Scatt.}} &
\multicolumn{4}{c}{\textbf{Temp.\ Global}} \\
\cmidrule(lr){2-5} \cmidrule(lr){6-9} \cmidrule(lr){10-12} \cmidrule(lr){13-15} \cmidrule(lr){16-19} \cmidrule(lr){20-23}
&
\rotatebox{90}{MRR@1} & \rotatebox{90}{MRR@3} & \rotatebox{90}{R@3} & \rotatebox{90}{Acc} &
\rotatebox{90}{MRR@1} & \rotatebox{90}{MRR@3} & \rotatebox{90}{R@3} & \rotatebox{90}{Acc} &
\rotatebox{90}{MRR@1} & \rotatebox{90}{MRR@3} & \rotatebox{90}{R@3} &
\rotatebox{90}{MRR@1} & \rotatebox{90}{MRR@3} & \rotatebox{90}{R@3} &
\rotatebox{90}{MRR@1} & \rotatebox{90}{MRR@3} & \rotatebox{90}{R@3} & \rotatebox{90}{Acc} &
\rotatebox{90}{MRR@1} & \rotatebox{90}{MRR@3} & \rotatebox{90}{R@3} & \rotatebox{90}{Acc} \\
\midrule[\heavyrulewidth]
VL-MemKnG (Ours)
& 0.59 & 0.62 & 51 & 59
& \third{0.57} & 0.62 & \third{59} & \third{61}
& \second{0.79} & \third{0.81} & \third{71}
& \best{0.72} & \third{0.74} & \third{54}
& \third{0.55} & \third{0.58} & 47 & \best{73}
& \best{0.72} & \best{0.73} & \second{57} & \second{73} \\
\midrule[\heavyrulewidth]
VL-KnG (GER-L)
& \second{0.65} & \third{0.72} & \second{61} & 54
& 0.51 & 0.60 & \second{60} & 49
& 0.67 & 0.77 & \second{77}
& 0.64 & 0.73 & \best{61}
& 0.28 & 0.29 & 23 & 43
& \third{0.42} & 0.48 & 46 & \third{54} \\
VL-KnG (GER-G)
& \best{0.68} & \best{0.74} & \best{66} & 59
& 0.53 & 0.63 & \best{65} & 47
& 0.72 & 0.78 & \third{74}
& \second{0.70} & \best{0.76} & \best{64}
& 0.32 & 0.33 & 26 & 43
& 0.41 & 0.43 & 36 & 46 \\
\midrule
Qwen 3.5+
& \second{0.65} & 0.71 & \second{61} & \best{70}
& 0.55 & \third{0.65} & \third{59} & \second{63}
& \best{0.81} & \best{0.89} & \best{81}
& 0.62 & 0.70 & 53
& \second{0.57} & \second{0.59} & \second{54} & \third{67}
& \third{0.42} & 0.45 & 38 & 41 \\
\midrule
Gemini 2.5 Flash
& 0.57 & 0.67 & 51 & \third{62}
& \second{0.59} & \second{0.69} & 57 & \best{69}
& 0.65 & 0.77 & 62
& 0.64 & 0.69 & 43
& 0.53 & 0.57 & \third{52} & \third{67}
& \second{0.50} & \third{0.56} & \second{53} & \third{70} \\
Gemini 2.5 Pro
& \third{0.60} & \second{0.73} & \third{60} & \second{68}
& \best{0.69} & \best{0.72} & 52 & 57
& \third{0.77} & \second{0.87} & 65
& \third{0.66} & \best{0.76} & 50
& \best{0.67} & \best{0.72} & \best{67} & \best{73}
& \second{0.50} & \second{0.62} & \best{63} & \best{78} \\
\bottomrule
\end{tabular}
}
\end{table}

These results support the central motivation behind the proposed hybrid-memory design. The spatio-temporal knowledge graph captures structured relational and object-centric information, while segment-level memory preserves broader contextual and temporal continuity. Together, these complementary retrieval mechanisms enable more reliable aggregation of evidence distributed across long temporal horizons.

For object-centric categories such as \emph{object search} and \emph{action-place association}, graph-based methods remain highly competitive, since relevant evidence can often be localized through objects, attributes, and spatial anchors alone. For scene-description and \emph{spatial-relation} questions, full-context vision--language models frequently achieve strong answer accuracy, suggesting that direct multimodal reasoning over broader visual context remains beneficial for fine-grained interpretation.

Overall, the per-category analysis shows that VL-MemKnG does not replace graph-based retrieval; rather, it extends it with contextual temporal memory that improves evidence organization and retrieval across distributed temporal events.

\subsection{Efficiency Analysis}
\label{subsec:efficiency_analysis}

We also evaluate query-time efficiency, since long-horizon navigation settings often require multiple questions over the same trajectory.
Full-context VLMs reprocess the visual context for each query, while VL-MemKnG shifts expensive computation into an offline memory-construction stage.

\begin{figure}[h!]
\begin{center}
\includegraphics[width=\textwidth]{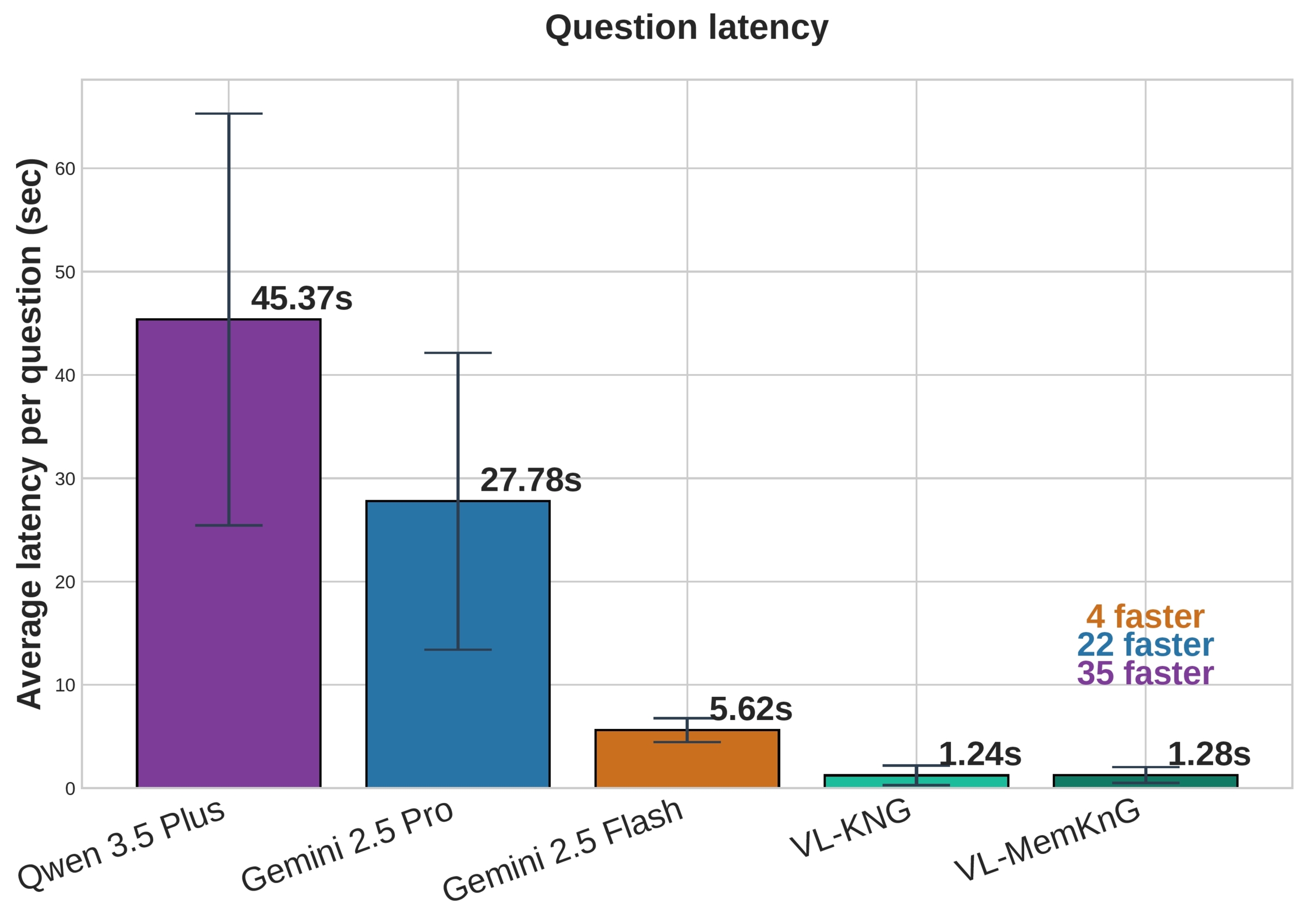}
\end{center}
\caption{Average question latency across methods. VL-MemKnG remains close to VL-KnG and is substantially faster than full-context VLM baselines.}
\label{fig:question_latency}
\end{figure}

Figure~\ref{fig:question_latency} compares average latency per question.
VL-MemKnG answers a question in approximately 1.28 seconds, close to VL-KnG at 1.24 seconds and substantially faster than full-context VLM baselines.
Gemini 2.5 Flash requires 5.62 seconds on average, Gemini 2.5 Pro requires 27.78 seconds, and Qwen 3.5+ requires 45.37 seconds.

Thus, VL-MemKnG is roughly $4\times$ faster than Gemini 2.5 Flash, $22\times$ faster than Gemini 2.5 Pro, and $35\times$ faster than Qwen 3.5+ at query time.

\begin{figure}[h!]
\begin{center}
\includegraphics[width=\textwidth]{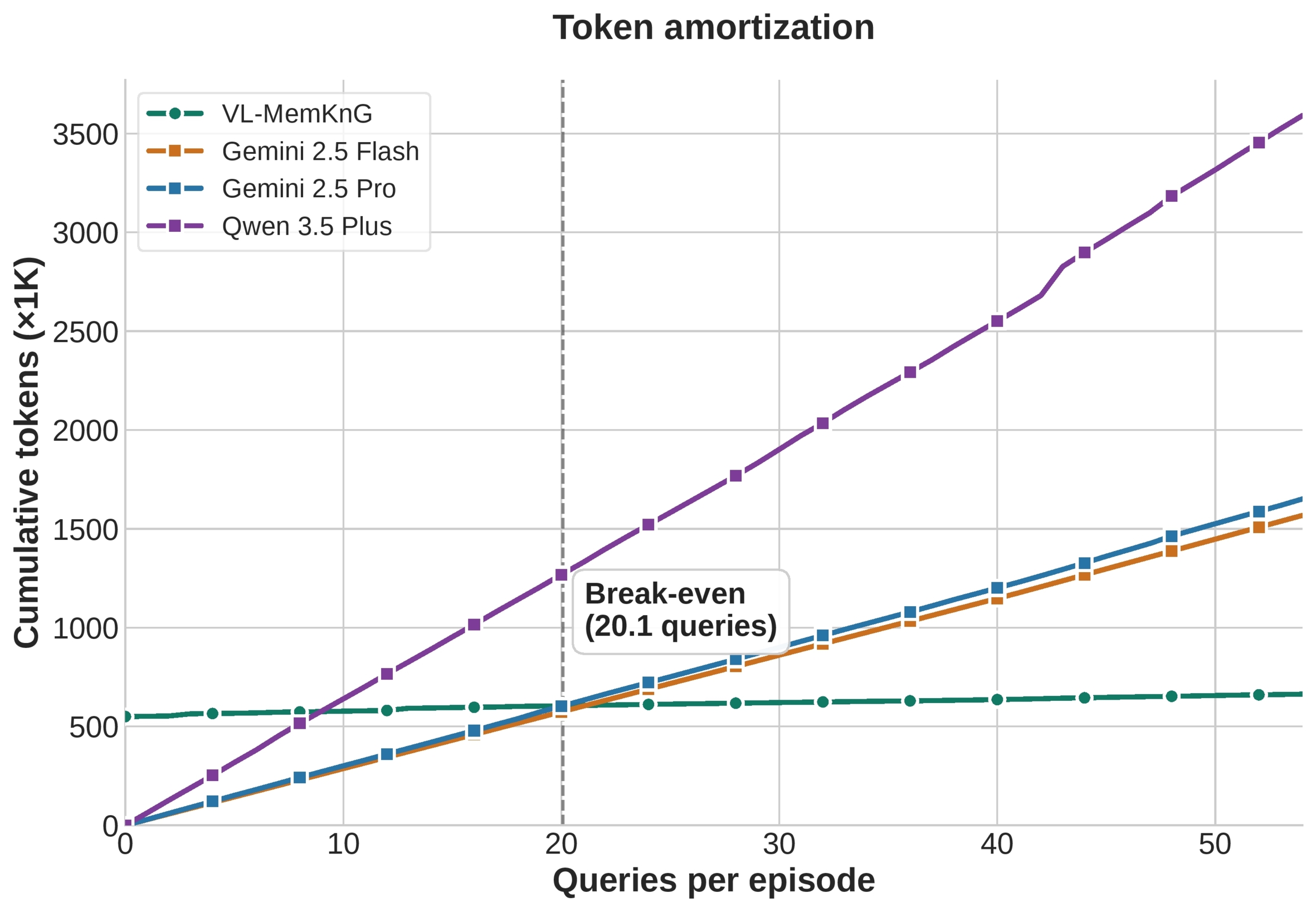}
\end{center}
\caption{Cumulative token cost as a function of the number of queries per episode. VL-MemKnG incurs an upfront memory-construction cost, but this cost is amortized over repeated queries, with break-even at approximately 20 queries.}
\label{fig:token_amortization}
\end{figure}

Figure~\ref{fig:token_amortization} shows cumulative token cost as the number of queries per episode increases.
VL-MemKnG starts with a non-zero offline cost due to memory construction, but this cost is amortized across repeated queries.
The break-even point occurs after approximately 20 queries, after which VL-MemKnG becomes more token-efficient than full-context VLM baselines.

These results highlight the practical advantage of the memory-centric design: VL-MemKnG is most beneficial when the same navigation trajectory is queried repeatedly.

\section{Ablation Studies}
\label{sec:ablation}

This section presents additional ablation studies for VL-MemKnG.
The experiments isolate three design choices in the retrieval pipeline: multimodal fusion weighting, retrieval depth, and query formulation for segment-level retrieval.

\subsection{Multimodal Fusion Weights}
\label{subsec:ablation_fusion}

Segment-level retrieval combines visual and textual similarity scores with weights $w_{\mathrm{vis}}$ and $w_{\mathrm{text}}$, where $w_{\mathrm{vis}}+w_{\mathrm{text}}=1$. Table~\ref{tab:ablation_fusion_weights} reports the effect of different weighting configurations. Balanced fusion provides the most consistent retrieval performance. The setting $w_{\mathrm{vis}}=w_{\mathrm{text}}=0.5$ achieves the strongest results across retrieval accuracy, recall, precision, and MRR, while also tying for the highest answer accuracy.

\begin{table}[htbp]
\centering
\caption{Ablation study of VL-MemKnG under different visual and textual weights for multimodal segment retrieval. The experiment uses $k_{\mathrm{KG}}=10$ and $k_{\mathrm{mem}}=10$ on two WalkieKnowledgeT+ trajectories with $84$ retrieval questions and $39$ answer-scored questions. Retrieval accuracy, recall@$k$, precision@$k$, and answer accuracy are reported as percentages; MRR@$k$ is unitless.}
\label{tab:ablation_fusion_weights}
\scriptsize
\setlength{\tabcolsep}{2pt}
\renewcommand{\arraystretch}{1.05}
\resizebox{\textwidth}{!}{%
\begin{tabular}{@{}cc*{12}{c}c@{}}
\toprule
\multirow{2}{*}{\textbf{$w_{\mathrm{vis}}$}} &
\multirow{2}{*}{\textbf{$w_{\mathrm{text}}$}} &
\multicolumn{3}{c}{\textbf{Retr.\ Acc.\ $\uparrow$}} &
\multicolumn{3}{c}{\textbf{Recall $\uparrow$}} &
\multicolumn{3}{c}{\textbf{Precision $\uparrow$}} &
\multicolumn{3}{c}{\textbf{MRR $\uparrow$}} &
\multirow{2}{*}{\parbox{2.1em}{\centering\textbf{Ans.}\\\textbf{Acc.\ $\uparrow$}}} \\
\cmidrule(lr){3-5} \cmidrule(lr){6-8} \cmidrule(lr){9-11} \cmidrule(lr){12-14}
& &
{@1} & {@3} & {@5} &
{@1} & {@3} & {@5} &
{@1} & {@3} & {@5} &
{@1} & {@3} & {@5} & \\
\midrule
$0$ & $1$ &
56 & 69 & 71 &
30 & 51 & 55 &
56 & 38 & 27 &
0.56 & 0.62 & 0.62 &
49 \\
$0.1$ & $0.9$ &
52 & 70 & 73 &
29 & 49 & 55 &
52 & 38 & 27 &
0.52 & 0.60 & 0.60 &
49 \\
$0.2$ & $0.8$ &
55 & 68 & 73 &
29 & 49 & 55 &
55 & 39 & 28 &
0.55 & 0.61 & 0.62 &
51 \\
$0.3$ & $0.7$ &
58 & 71 & 74 &
30 & 53 & 56 &
58 & 41 & 28 &
0.58 & 0.64 & 0.64 &
49 \\
\textbf{$0.5$} & \textbf{$0.5$} &
\textbf{60} & \textbf{74} & \textbf{77} &
\textbf{33} & \textbf{55} & \textbf{59} &
\textbf{60} & \textbf{42} & \textbf{29} &
\textbf{0.60} & \textbf{0.65} & \textbf{0.66} &
\textbf{51} \\
$0.7$ & $0.3$ &
56 & 68 & 73 &
30 & 52 & 57 &
56 & 40 & 28 &
0.56 & 0.61 & 0.62 &
46 \\
$0.9$ & $0.1$ &
52 & 64 & 70 &
27 & 47 & 50 &
52 & 35 & 24 &
0.52 & 0.58 & 0.59 &
38 \\
\bottomrule
\end{tabular}%
}%
\end{table}

Text-only retrieval remains competitive, suggesting that segment captions encode strong semantic information, but it underperforms balanced fusion on Top-1 retrieval and ranking quality. In contrast, visual-dominated weighting degrades performance, indicating that visual similarity alone is insufficient without caption-aligned semantics. We therefore use balanced fusion in the main experiments.

\subsection{Retrieval Depth}
\label{subsec:ablation_kg_mem_topk}

We next examine how many candidates should be retained from each retrieval stream before reasoning. Table~\ref{tab:ablation_kg_cap_topk} reports results for different retrieval-depth configurations. The method is relatively insensitive to moderate changes in retrieval depth once both retrieval streams provide enough candidates. The two strongest configurations, $(k_{\mathrm{KG}},k_{\mathrm{mem}})=(7,7)$ and $(10,10)$, achieve very similar results. The $(7,7)$ setting gives slightly stronger early-ranking metrics, while $(10,10)$ provides broader evidence coverage at Top-5. This suggests that increasing retrieval depth mainly improves evidence diversity rather than Top-1 ranking quality. Given the small performance gap, we use $(10,10)$ in the main experiments to preserve a richer candidate pool for the reasoning stage.

\begin{table}[htbp]
\centering
\caption{
Ablation study of VL-MemKnG under different retrieval depths with $w_{\mathrm{vis}}=w_{\mathrm{text}}=0.5$.The experiment uses two WalkieKnowledgeT+ trajectories with $84$ questions. $k_{\mathrm{KG}}$ denotes the top-$k$ cutoff for knowledge-graph retrieval, and $k_{\mathrm{mem}}$ denotes the top-$k$ cutoff for segment-level memory retrieval. Retrieval accuracy, recall@$k$, precision@$k$, and answer accuracy are reported as percentages; MRR@$k$ is unitless.
}
\label{tab:ablation_kg_cap_topk}
\scriptsize
\setlength{\tabcolsep}{2pt}
\renewcommand{\arraystretch}{1.05}
\resizebox{\textwidth}{!}{%
\begin{tabular}{@{}cc*{12}{c}c@{}}
\toprule
\multirow{2}{*}{\textbf{$k_{\mathrm{KG}}$}} &
\multirow{2}{*}{\textbf{$k_{\mathrm{mem}}$}} &
\multicolumn{3}{c}{\textbf{Retr.\ Acc.\ $\uparrow$}} &
\multicolumn{3}{c}{\textbf{Recall $\uparrow$}} &
\multicolumn{3}{c}{\textbf{Precision $\uparrow$}} &
\multicolumn{3}{c}{\textbf{MRR $\uparrow$}} &
\multirow{2}{*}{\parbox{2.1em}{\centering\textbf{Ans.}\\\textbf{Acc.\ $\uparrow$}}} \\
\cmidrule(lr){3-5} \cmidrule(lr){6-8} \cmidrule(lr){9-11} \cmidrule(lr){12-14}
& &
{@1} & {@3} & {@5} &
{@1} & {@3} & {@5} &
{@1} & {@3} & {@5} &
{@1} & {@3} & {@5} & \\
\midrule
$5$ & $7$ &
57 & 71 & 77 &
31 & 54 & 59 &
57 & 41 & 29 &
0.57 & 0.63 & 0.65 &
49 \\
$7$ & $5$ &
60 & 71 & 76 &
33 & 55 & 59 &
60 & 42 & 29 &
0.60 & 0.65 & 0.66 &
54 \\
\textbf{$7$} & \textbf{$7$} &
\textbf{62} & 71 & 75 &
\textbf{33} & 54 & 58 &
\textbf{62} & 41 & 28 &
\textbf{0.62} & \textbf{0.66} & \textbf{0.67} &
51 \\
\textbf{$10$} & \textbf{$10$} &
\textbf{61} & 71 & \textbf{77} &
\textbf{34} & 54 & \textbf{60} &
\textbf{61} & 42 & \textbf{30} &
\textbf{0.61} & 0.65 & \textbf{0.67} &
51 \\
\bottomrule
\end{tabular}%
}%
\end{table}

\subsection{Query Formulation}
\label{subsec:ablation_query_embed}

Finally, we study how the textual query formulation used for segment-level retrieval affects performance. Table~\ref{tab:ablation_query_embedding} compares embedding the full natural-language question against embedding the transformed object-centric retrieval query. Using the transformed retrieval query improves Top-3 and Top-5 retrieval accuracy, recall, precision, and MRR, while Top-1 metrics remain unchanged. This suggests that retrieval-focused object-centric queries better align the embedding with visually grounded content. By contrast, embedding the full question yields higher answer accuracy, indicating that additional linguistic context can help the reasoning stage select the correct answer. Since the transformed retrieval query provides stronger retrieval performance overall, we use it in the main experiments.

\begin{table}[htbp]
\centering
\caption{
Ablation study of VL-MemKnG using either the full question or the extracted retrieval query as text input for segment embedding search, with $w_{\mathrm{vis}}=w_{\mathrm{text}}=0.5$ and $(k_{\mathrm{KG}},k_{\mathrm{mem}})=(10,10)$. The experiment uses two WalkieKnowledgeT+ trajectories with $84$ questions. Retrieval accuracy, recall@$k$, precision@$k$, and answer accuracy are reported as percentages; MRR@$k$ is unitless.
}
\label{tab:ablation_query_embedding}
\scriptsize
\setlength{\tabcolsep}{2pt}
\renewcommand{\arraystretch}{1.05}
\resizebox{\textwidth}{!}{%
\begin{tabular}{@{}l*{12}{c}c@{}}
\toprule
\multirow{2}{*}{\textbf{Embedding text}} &
\multicolumn{3}{c}{\textbf{Retr.\ Acc.\ $\uparrow$}} &
\multicolumn{3}{c}{\textbf{Recall $\uparrow$}} &
\multicolumn{3}{c}{\textbf{Precision $\uparrow$}} &
\multicolumn{3}{c}{\textbf{MRR $\uparrow$}} &
\multirow{2}{*}{\parbox{2.1em}{\centering\textbf{Ans.}\\\textbf{Acc.\ $\uparrow$}}} \\
\cmidrule(lr){2-4} \cmidrule(lr){5-7} \cmidrule(lr){8-10} \cmidrule(lr){11-13}
& {@1} & {@3} & {@5} & {@1} & {@3} & {@5} & {@1} & {@3} & {@5} & {@1} & {@3} & {@5} & \\
\midrule
Object-centric query &
61 & \textbf{71} & \textbf{77} &
\textbf{34} & \textbf{54} & \textbf{60} &
61 & \textbf{42} & \textbf{30} &
0.61 & \textbf{0.65} & \textbf{0.67} &
49 \\
Full question &
61 & 68 & 75 &
33 & 49 & 56 &
61 & 36 & 26 &
0.61 & 0.64 & 0.65 &
54 \\
\bottomrule
\end{tabular}%
}%
\end{table}

\section{Discussion}
\label{sec:discussion}

\subsection{Comparison of Retrieval Approaches}

Unlike full-context vision--language models that repeatedly process the entire video during inference, VL-MemKnG maintains persistent memory representations that support efficient retrieval and reasoning over long egocentric trajectories without repeated full-video processing. The experimental results highlight complementary strengths and weaknesses across existing approaches for long-video question answering. Full-context vision--language models achieve strong answer accuracy and high-$k$ retrieval performance by directly accessing the entire visual context at inference time. However, this requires repeated processing of long videos for each query and results in substantially higher latency and token consumption.

Graph-based retrieval methods such as VL-KnG remain effective for object-centric localization and spatial reasoning, where relevant evidence can often be recovered through localized object relations and structured graph traversal. At the same time, their performance degrades on temporally distributed questions that require evidence aggregation across multiple non-cooccurring moments.

VL-MemKnG occupies a complementary position between these paradigms. By combining spatio-temporal knowledge graph retrieval with segment-level contextual memory, the framework improves evidence ranking while avoiding repeated full-video processing at query time.

\subsection{Temporal Reasoning and Hybrid Memory}

The strongest gains from VL-MemKnG appear on temporal-global and temporally scattered aggregation questions introduced in WalkieKnowledgeT+. These question types require the system not only to localize relevant evidence, but also to organize evidence distributed across distant temporal intervals.

The per-category analysis suggests that segment-level memory complements the spatio-temporal knowledge graph by providing temporally indexed contextual representations beyond localized relational structure. In particular, segment-level memory helps preserve contextual information that is difficult to recover through graph-local relations alone, improving evidence organization for chronology-aware questions spanning multiple temporal moments.

\subsection{Efficiency and Scalability}

From an efficiency perspective, VL-MemKnG shifts most computational cost to an offline memory-construction and indexing stage. Query-time reasoning therefore operates over compact memory representations rather than full-video inputs, resulting in substantially lower latency compared with full-context VLM baselines.

The efficiency analysis further shows that the offline memory cost becomes amortized as the number of queries increases. This behavior is particularly important for long-horizon navigation settings, where the same trajectory may be queried repeatedly for different tasks or reasoning objectives. Under these conditions, persistent memory representations provide a practical alternative to repeated full-context inference.

\subsection{Limitations and Future Work}

Despite these advantages, the proposed framework has several limitations.

First, the framework relies on caption-derived memory representations and therefore inherits the limitations of the underlying vision--language model. The quality of automatically generated captions directly affects both segment-level memory retrieval and knowledge-graph construction, and errors or omissions in caption generation may propagate into downstream retrieval and reasoning stages.

Second, the knowledge graph is currently constructed offline and does not support incremental updates, limiting applicability in continuously evolving environments or streaming settings. Future work could investigate dynamic graph updates and online memory refinement.

\subsection{Summary}

Overall, the results demonstrate that combining spatio-temporal knowledge graphs with segment-level contextual memory provides an effective framework for long-horizon video question answering. The hybrid-memory design improves retrieval performance on temporally structured reasoning tasks while maintaining efficient query-time inference through persistent memory representations. These findings suggest that memory-centric retrieval architectures offer a practical alternative to repeated full-context video processing for long egocentric navigation trajectories.












\section*{Data Availability Statement}
The raw data supporting the conclusions of this article will be made available by the authors upon request. The source code and implementation details necessary to reproduce the reported experiments will be made publicly available upon publication.


\bibliographystyle{plainnat} 
\raggedbottom
\bibliography{test}

\end{document}